# Exploring hyperelastic material model discovery for human brain cortex: multivariate analysis vs. artificial neural network approaches

Jixin Hou[1], Nicholas Filla[1], Xianyan Chen[2], Mir Jalil Razavi[3], Tianming Liu[4], Xianqiao Wang[1]*

[1] School of ECAM, College of Engineering, University of Georgia, Athens, GA, 30602

[2] Department of Statistics, University of Georgia, Athens, GA, 30602

[3] Department of Mechanical Engineering, Binghamton University, Binghamton, NY, 13902

[4] School of Computing, University of Georgia, Athens, GA, 30602

*Corresponding author: xqwang@uga.edu

## Abstract

The intricate architecture of the human brain exhibits complex mechanical properties and endows it to perform pivotal functions. Traditional computational methods, such as the finite element analysis, have provided valuable insights into uncovering the underlying mechanisms of brain physical behaviors. However, precise predictions of brain physics require effective constitutive models to represent the intricate mechanical properties of brain tissue. In this study, we aimed to identify the most favorable constitutive material model for human brain tissue. To achieve this, we applied artificial neural network and multiple regression methods to a generalization of widely accepted classic models, and compared the results obtained from these two approaches. To evaluate the applicability and efficacy of the model, all setups were kept consistent across both methods, except for the approach to prevent potential overfitting. Our results demonstrate that artificial neural networks are capable of automatically identifying accurate constitutive models from given admissible estimators. Nonetheless, the five-term and two-term neural network models trained under single-mode and multi-mode loading scenarios, were found to be suboptimal and could be further simplified into two-term and single-term, respectively, with higher accuracy using multiple regression. Our findings highlight the importance of hyperparameters for the artificial neural network and emphasize the necessity for detailed cross-validations of regularization parameters to ensure optimal selection at a global level in the development of material constitutive models. This study validates the applicability and accuracy of artificial neural network to automatically discover constitutive material models with proper regularization as well as the benefits in model simplification without compromising accuracy for traditional multivariable regression.

# 1 Introduction

The human brain tissue possesses an intricate architecture that enables it to operate as the pivotal regulator of numerous essential physiological processes within the body. The complex structure of the brain exhibits sophisticated mechanical behaviors such as heterogeneity, extreme compliance, high fragility, biphasic properties, and ultrasoft nature [1]. Increasing studies have demonstrated that mechanics play a critical role in modulating brain function and dysfunction, including brain development [2-4], degradation [5-7], disease progression [8, 9], and traumatic brain injury [10, 11]. To unravel the underlying mechanisms of these complex processes, computational simulations, including finite element method, have emerged as a potent and versatile tool with diverse applications [12-16]. Nonetheless, to ensure accurate computational predictions, it is crucial to have effective constitutive models that can adeptly capture the intricate mechanical properties of the brain tissue.

An important application of constitutive models is to characterize the time-independent response of the brain tissue under finite deformations, where the tissue is typically modeled as a hyperelastic, rubber-like substance [17]. Therefore, the classic invariant-based models developed for rubbers, such as the neo-Hookean and Mooney Rivlin models, have been utilized to calibrate the elasticity of the brain tissue [18-20]. However, the straightforward combination of linear invariant-form terms fails to capture the nonlinear mechanics of the brain tissue [21]. To enforce the nonlinearity, transformations incorporating exponential, logarithmic, polynomial, or coupling forms have been integrated into constitutive models, such as the Yeoh model [22, 23], Arruda-Boyce model [22, 24], Gent model [25], and Holzapfel model [26]. Despite the successful application of these nonlinear constitutive models in various aspects, a consensus on their superiority remains elusive [27]. Furthermore, most models are calibrated using single-mode data, either tension, compression, or shear. Duo to the complex mechanical behavior of the brain tissue, models trained solely on single-mode data are inadequate for capturing the constitutive behavior of brain tissue under arbitrary loading conditions, which emphases the necessity for models trained on multiple loading modes data simultaneously [21, 28]. Furthermore, the pronounced conditioning effect results in significant variations in the range of experimental data both quantitatively and qualitatively, making it challenging to determine competing constitutive models for the human brain tissue[13, 29].

Data-driven techniques, particularly neural networks, have shown great promise in dealing with the intricate nonlinear relationships present in complex datasets, and have emerged as a valuable tool to facilitate the discovery of constitutive models. Initially, classic neural networks with multiple fully connected hidden layers were employed for model discovery. Trained with massive experimental strain and stress data as input and output, respectively, these networks are capable of accurately interpolating experimental data even without a thorough understanding of the underlying physics. However, classic neural networks struggle with extrapolation and fail to achieve satisfactory predictive performance on test data. Additionally, due to their black-box nature, the neural network algorithms lack transparency which makes it challenging to interpret the models [30, 31]. To address these limitations, physics-informed neural networks (PINN) have been introduced by incorporating underlying physical laws or constraints into the neural network architecture. This approach improves both model accuracy and interpretability by modifying objective functions to include known physical laws, such as partial differential equations, and prescribed constraints, allowing it to solve physics-based problems while satisfying certain constraints [32-34]. Similarly, constitutive artificial neural networks explicitly incorporate constitutive equations and corresponding constitutive constraints, such as thermodynamics, polyconvexity and symmetry, into the network architecture design [35]. These artificial neural networks have been successfully utilized to predict the constitutive behavior of rubber-like materials [36-38], and soft tissues [39-42]. While effective, the integration of constitutive constrains and the detailed physical meanings of neural network parameters in neural network still lack intuitive interpretation.

In a recent development, Linka et al. [43] have introduced a novel family of neural networks that reverse-engineer the network structures from a generalization of widely accepted constitutive models, including the neo-Hookean, Mooney Rivlin, Demiray, Gent and Holzapfel models. The proposed networks incorporate rigorous constitutive equations and well-defined thermodynamic constraints into their structures to ensure robustness. Moreover, the network parameters, particularly the weightings, have clear physical interpretations that correspond to material parameters, such as stiffness. With this effective strategy, they successfully obtained a series of constitutive models that accurately predicted the complex behavior of soft biological tissues with satisfying accuracy, including rabbit skin [44], and human brain tissue [45]. However, while this method was capable of automatically discovering models for brain tissues, the discovered models were not in a concise format like classic constitutive models, especially for the model derived from single-mode training scenario, which consists of more than 10 terms. Naturally, this raises the question of whether the discovered model is globally optimal and whether it possesses adequate accuracy with forms as simple as possible.

In the present study, we aim to discover the constitutive model for the human brain tissue using traditional statistical methods, specifically multivariant regression. To achieve this, we followed the same setups as the artificial neural network, by regressing our model with uniaxial tension, compression, and simple shear experimental data from the human brain cortex [21]. We performed two distinct training scenarios (i.e., single-mode and multi-mode training) and determined the optimal model using the Bayesian information criterion (BIC). The main objective of this study is to find the optimal model that can accurately capture the complex mechanical behavior of the human brain cortex and to compare the applicability and efficiency of neural network and multiple regression methods in constitutive model discovery.

The manuscript is structured as follows: In Section 2, we provide an overview of the fundamentals of continuum mechanics, along with a brief description of the algorithms used in both multiple regression and neural network methods. Section 3 presents the models discovered by both methods, followed by a comparison of their applicability and efficiency in Section 4. Finally, the findings are summarized in the form of a conclusion in Section 5.

## 2 Theoretical Method, Statistical Analysis, and Neural Network Algorithm

### 2.1 Constitutive modeling

**Kinematics and stress**

In the realm of continuum mechanics, we characterize the kinematics of deformation with deformation map $x = \varphi(X)$, which maps a material particle from original position $X$ in the undeformed (reference) configuration $\mathcal{B}_0$ to its new position $x$ in the deformed (spatial) configuration $\mathcal{B}_t$. In addition, we introduce the deformation gradient $F = \nabla_X \varphi$ to measure the mapping of the line element from reference to spatial configuration, and the Jacobian $J = \det F$ to describe local volume alternation from reference to spatial configuration. For any physically admissible deformation, the Jacobian $J$ will always be positive, if $0 < J < 1$ the deformation is a contraction, while $1 < J$ denotes a dilation. Left multiplying $F$ with its transpose $F^T$ produces the right Cauchy Green deformation tensor $C = F^T F$, which is advantageous to eliminates the translational and rotational component present in deformation gradient due to its symmetric nature. In the undeformed state, both deformation gradient and Cauchy Green deformation tensor are identical to the unit tensor, $F = I$, $C = I$, and Jacobian equals one, $J = 1$. Furthermore, we introduce two types of stress: the symmetric Cauchy stress $\sigma$, defined as the force per deformed area, and asymmetric first Piola-Kirchhoff stress $P$, defined as the force per undeformed area. The transpose of the latter one is also known as nominal stress and is commonly measured directly from experiments. The relation between these two stresses is described by Piola transformation:

$$\boldsymbol{P} = J\boldsymbol{\sigma}\boldsymbol{F}^{-T} \quad (1)$$

**Constitutive equations**

The constitutive equation describes internal mechanical response of materials under external stimuli by constructing a quantitative linkage between the measure of stress state and deformation, for example the first Piola-Kirchhoff stress and deformation gradient, $\boldsymbol{P} = \boldsymbol{P}(\boldsymbol{F})$. For hyperelastic materials, the constitutive relations can be described by postulating the existence of a Helmholtz free-energy function, $\Psi(\boldsymbol{F})$, which is a function of nine-component tensor $\boldsymbol{F}$. Since $\Psi$ primarily depends on stretching, rather than rotation or translation, it can be expressed as a function of six-component tensor $\boldsymbol{C}$, namely $\Psi(\boldsymbol{F}) = \Psi(\boldsymbol{C})$. The relation, complying with the second law of thermodynamics, can be further presented as follows,

$$\boldsymbol{P} = \frac{\partial \Psi}{\partial \boldsymbol{F}} - p\boldsymbol{F}^{-T} \quad (2)$$

where $p$ is the Lagrange multiplier and can be identified as hydrostatic pressure, the modified term $-p\boldsymbol{F}^{-T}$ is introduced to enforce perfect incompressibility.

For an isotropic material, its mechanical response is direction-independent, thus we can use three complete and irreducible invariants to describe its strain energy function,

$$\Psi(\boldsymbol{C}) = \Psi(I_1, I_2, I_3) \quad \text{with } I_1 = \text{tr}\boldsymbol{C}, I_2 = \text{tr}(\text{cof}\boldsymbol{C}), I_3 = \det \boldsymbol{C} \quad (3)$$

where $\text{cof}\boldsymbol{C} = \det(\boldsymbol{C})\boldsymbol{C}^{-T}$ denotes the cofactor of right Cauchy Green deformation tensor. In the undeformed state, $I_1 = I_2 = 3$, and under the incompressibility constraint, $I_3$ remains constant, $I_3 \equiv 1$. Therefore, Eq. 2 can be expanded with respect to Eq. 3 using chain rules,

$$\boldsymbol{P} = \frac{\partial \Psi(I_1, I_2, I_3)}{\partial \boldsymbol{F}} - p\boldsymbol{F}^{-T} = \frac{\partial \Psi}{\partial I_1}\frac{\partial I_1}{\partial \boldsymbol{F}} + \frac{\partial \Psi}{\partial I_2}\frac{\partial I_2}{\partial \boldsymbol{F}} - p\boldsymbol{F}^{-T} \quad (4)$$

where $\partial I_1/\partial \boldsymbol{F} = 2\boldsymbol{F}$, $\partial I_2/\partial \boldsymbol{F} = 2I_1\boldsymbol{F} - \boldsymbol{F}\boldsymbol{F}^T\boldsymbol{F}$

The utilization of an invariants-based strain energy function above empowers the computation of the first Piola Kirchhoff stress concerning the deformation gradient, which can be directly measured from experiments under certain circumstances, including uniaxial tension and compression, as well as simple shear.

In the case of unconfined uniaxial tension and compression test, the stretched specimen deforms uniformly in a single direction. Under the assumption of isotropy and perfect incompressibility, the deformation gradient $F$ and right Cauchy Green deformation tensor $C$ take on following matrix forms relative to the stretch $\lambda$,

$$[F] = \begin{bmatrix} \lambda & 0 & 0 \\ 0 & 1/\sqrt{\lambda} & 0 \\ 0 & 0 & 1/\sqrt{\lambda} \end{bmatrix} \qquad [C] = \begin{bmatrix} \lambda^2 & 0 & 0 \\ 0 & 1/\lambda & 0 \\ 0 & 0 & 1/\lambda \end{bmatrix} \qquad (5)$$

with three invariants

$$I_1 = \lambda^2 + \frac{2}{\lambda} \qquad I_2 = 2\lambda + \frac{1}{\lambda^2} \qquad I_3 = 1 \qquad (6)$$

Accordingly, the nominal uniaxial stress can be calculated using Eq. 4,

$$P_{11} = 2\left(\frac{\partial \Psi}{\partial I_1} + \frac{1}{\lambda}\frac{\partial \Psi}{\partial I_2}\right)\left(\lambda - \frac{1}{\lambda^2}\right) \qquad P_{22} = P_{33} = 0 \qquad (7)$$

Note that, hydrostatic pressure $p$ is determined from zero stress condition in the transverse directions for unconfined test.

For the simple shear test, a prescribed amount of shear $\gamma$ is applied to an isotropic and perfectly incompressible specimen. The resulting deformation gradient $F$ and right Cauchy Green deformation tensor $C$ take on following matrix form,

$$[F] = \begin{bmatrix} 1 & \gamma & 0 \\ 0 & 1 & 0 \\ 0 & 0 & 1 \end{bmatrix} \qquad [C] = \begin{bmatrix} 1 & \gamma & 0 \\ \gamma & 1+\gamma^2 & 0 \\ 0 & 0 & 1 \end{bmatrix} \qquad (8)$$

with three invariants,

$$I_1 = 3 + \gamma^2 \qquad I_2 = 3 + \gamma^2 \qquad I_3 = 1 \qquad (9)$$

The nominal shear stress $P_{12}$ is independent with hydrostatic pressure $p$, resulting in following explicit form,

$$P_{12} = 2\left(\frac{\partial \Psi}{\partial I_1} + \frac{\partial \Psi}{\partial I_2}\right)\gamma \qquad (10)$$

**Strain energy function**

Given the deformation gradient and energy function, the nominal tensile and compressive stress,

as well as shear stress, can be calculated using Eq. 7 and 10 for certain special cases. However, in order to ensure the admissibility of strain energy function in physics, several conditions must be fulfilled. These conditions are briefly discussed below, with a more detailed introduction available in [43].

First, the energy function $\Psi$ must be objective, meaning that it should not depend on the choice of material frame. This requires the equality $\Psi(\boldsymbol{QF}) = \Psi(\boldsymbol{F})$ to be satisfied for any deformation gradient $\boldsymbol{F}$ and orthogonal tensor $\boldsymbol{Q}$.

Second, the energy function $\Psi$ should reflect material symmetry, ensuring that the constitutive response remains consistent under transformations of the reference configuration. In this manuscript, the isotropy symmetry is considered to constrain the unknowns of energy function to three invariants, which significantly simplifies the input for strain energy function prediction in both neural network and multiple regression.

Third, the energy function $\Psi$ must comply with certain physical restrictions: (I) $\Psi$ should be non-negative for any deformation gradient, $\Psi(\boldsymbol{F}) \geq 0$; (II) $\Psi$ should vanish in the reference configuration, $\Psi(\boldsymbol{I}) = 0$, which is also referred to as normalization condition; (III) $\Psi$ should tend to $+\infty$ if either $J$ approaches $+\infty$ or $0^+$, which is known as growth condition for hyperelastic constitutive models.

Furthermore, the energy function $\Psi$ needs to be polyconvex, which means its construction is convex in terms of deformation gradient, as well as its cofactor and determinant [46],

$$\Psi(\boldsymbol{F}) = P(\boldsymbol{F}, \text{cof}\boldsymbol{F}, \det\boldsymbol{F}) \tag{11}$$

Polyconvexity implies ellipticity and ensures material stability of a constitutive model, by constraining the shape of the models to be convex [47]. Moreover, polyconvexity guarantees that the energy function attains its global minimum only at thermodynamic equilibrium in the reference configuration. To ensure polyconvexity, every step in constructing the energy function must preserve polyconvexity [35]. Consistent with previous work [48], we impose this condition with the additive combination of polyconvex invariants, since addictive manipulation preserves convexity,

$$\Psi(\boldsymbol{F}) = \Psi(I_1, I_2, I_3) = \Psi_1(I_1) + \Psi_2(I_2) + \Psi_3(I_3) \tag{12}$$

here $\Psi_1$, $\Psi_2$, $\Psi_3$ can be chosen from any polyconvex subfunctions. For this study, we selected subfunctions with forms similar to well-known constitutive models, such as the neo-Hookean model $\Psi^{\text{N-H}}$ [49], the two-terms Mooney Rivlin model $\Psi^{\text{M-R}}$ [50], the Demiray model $\Psi^{\text{Dmry}}$ [51], and the Gent models $\Psi^{\text{Gent}}$ [52].

$$\Psi^{\text{N-H}} = \frac{1}{2}\mu(I_1 - 3) \tag{13}$$

$$\Psi^{\text{M-R}} = \left(\frac{1}{2}\mu - C_2\right)(I_1 - 3) + C_2(I_2 - 3) \tag{14}$$

$$\Psi^{\text{Dmry}} = \frac{1}{2}\mu \left(\exp[\beta(I_1 - 3)] - 1\right)/\beta \tag{15}$$

$$\Psi^{\text{Gent}} = \frac{1}{2}\mu\eta \ln(1 - (I_1 - 3)/\eta) \tag{16}$$

These models have been demonstrated to be physically meaningful and appropriate for characterizing the mechanical response of brain tissue [53]. In application, we have selected three distinct subfunctions: linear, exponential, and logarithmic, which also serve as the activate function for the neural network. These functions are uniformly applied to $\Psi_1$, $\Psi_2$, $\Psi_3$, with $I_1$, $I_2$ as input arguments. Notably, we only consider $I_1$, $I_2$ as our arguments as $I_3$ remains constant at 1 due to the incompressibility condition. Following the normalization condition, both $I_1$ and $I_2$ are subtracted by 3 to maintain the stress-free state in the reference configuration. In addition, linear and quadratic forms are introduced for each modified argument to enrich model selection, resulting in four different inputs: $[I_1 - 3]$, $[I_2 - 3]$, $[I_1 - 3]^2$, $[I_2 - 3]^2$. Ultimately, a total of 12 options have been made available for selecting the optimal hyperelastic model, which rigorously satisfies conditions of objectivity, material symmetry, physical restrictions and polyconvexity by design.

In this manuscript, our objective is to identify the optimal hyperelastic model that can accurately capture the nonlinear mechanical behavior of heterogeneous brain tissue, such as the cortex. To achieve this, we employed both statistical and neural network methodologies. Two distinct approaches were utilized to enhance the model selection process, namely multiple regression, and constitutive artificial neural network, with their algorithms schematically illustrated in Figure 1. More detailed descriptions of the methods are presented in subsequent sections.

## 2.2 Multiple regression

Multiple regression is a statistical technique that assesses the association between multiple independent variables and a dependent variable. In our study, we employed this approach to identify the optimal energy function $\Psi$ from a pool of 12 potential options. A representative example including all the available selections is presented below,

$$\begin{aligned}\Psi(I_1, I_2) = &\ b_1[I_1 - 3] + b_5(\exp(b_{14}[I_1 - 3]) - 1) - b_9 (\ln(1 - b_{17}[I_1 - 3])) \\ &+ b_2[I_2 - 3] + b_6(\exp(b_{15}[I_2 - 3]) - 1) - b_{10}(\ln(1 - b_{18}[I_2 - 3])) \\ &+ b_3[I_1 - 3]^2 + b_7(\exp(b_{16}[I_1 - 3]^2) - 1) - b_{11}(\ln(1 - b_{19}[I_1 - 3]^2)) \\ &+ b_4[I_2 - 3]^2 + b_8(\exp(b_{17}[I_2 - 3]^2) - 1) - b_{12}(\ln(1 - b_{20}[I_2 - 3]^2))\end{aligned} \quad (17)$$

where $b$'s are the unknown weightings with a dimension size of 20, which are determined from calibrations. Although the model may achieve satisfactory performance, its complexity makes it difficult to interpret, and predictions may experience high variance due to overfitting. In order to mitigate the risk of overfitting, we introduced the Bayesian information criterion (BIC) [54, 55],

$$BIC = N \ln\left(\frac{RSS}{N}\right) + \ln(N)m \quad (18)$$

where $N$ is the number of data points in samples, or equivalently the sample size, $m$ is the number of free parameters to be estimate, and $RSS = \sum_{i=1}^{N}(y_i - \hat{y}_i)^2$ is the residual sum of squares, $y_i$ and $\hat{y}_i$ represent the $i$th variable in sample and prediction, respectively. The former term in BIC represents the loss of model fitting, while the latter term denotes the penalty of model complexity. Therefore, BIC measures the balance in trade-off between model accuracy and complexity.

To determine the optimal strain energy function from the 12 available options, we utilized the best subset approach. By listing and comparing all potential models, the optimal model was determined based on the lowest BIC value. The multiple regression scheme is illustrated in Figure 1b. We started by randomly selecting one term, resulting in 12 possible models, and then used the nonlinear curve fitting function *lsqnonlin* in MATLAB to calibrate the weightings. Note that, the data for calibration is nominal stress, which can be calculated from energy function using Eq. 2 or 4. During calibration, the least-squares algorithm based on mean absolute percentage error (MAPE) loss was used,

$$\mathcal{L}(\boldsymbol{b};\boldsymbol{F}) = \frac{100}{N}\sum_{i=1}^{N}\left\|\frac{\boldsymbol{P}(\boldsymbol{F}_i) - \widehat{\boldsymbol{P}}_i}{\boldsymbol{P}(\boldsymbol{F}_i)}\right\|_1 \tag{19}$$

where $\widehat{\boldsymbol{P}}$ is the prediction of the first Piola-Kirchhoff stress derived from the regression model, $\boldsymbol{P}$ is stress from experimental data, and $\|\ \|_1$ is L1 norm or absolute value of a vector. The goodness of fit is measured with R-squared value, $R^2 = 1 - \sum_{i=1}^{N}(\boldsymbol{P}_i - \widehat{\boldsymbol{P}}_i)^2 / \sum_{i=1}^{N}(\boldsymbol{P}_i - \overline{\boldsymbol{P}})^2$ where $\overline{\boldsymbol{P}}$ is the mean of experimental stress data. Subsequently, we repeated the aforementioned weighting calibration process for models with 2 terms, 3 terms, and up to 12 terms. At each step, we saved the optimal weightings $\boldsymbol{b}$ and corresponding local minimum BIC values. Finally, we compared these BIC values, and the optimal model was determined based on the global minimum BIC value.

## 2.3 Artificial neural network

A feedforward neural network is constructed based on the structure illustrated in Figure 1c. The neural network has four independent inputs derived from invariants, namely $[I_1 - 3]$, $[I_2 - 3]$, $[I_1 - 3]^2$, $[I_2 - 3]^2$, and an energy function $\Psi$ as the output. To capture the nonlinearity of the hyperelastic constitutive model, a hidden layer was introduced with three activate functions: linear, exponential, and logarithmic. The connections between these three layers are sparse rather than fully connected to enforce the polyconvexity. For a more comprehensive overview of this constitutive artificial neural network, readers are suggested to refer to [43]. Based on the neural network's structure, the energy function $\Psi$ can be explicitly expressed as fellows,

$$\begin{aligned}\Psi(I_1, I_2) &= w_{2,1}w_{1,1}\,[I_1 - 3] + w_{2,2}\left(\exp(w_{1,2}\,[I_1 - 3]) - 1\right) - w_{2,3}\left(\ln(1 - w_{1,3}\,[I_1 - 3])\right) \\ &+ w_{2,4}w_{1,4}\,[I_2 - 3] + w_{2,5}\left(\exp(w_{1,5}\,[I_2 - 3]) - 1\right) - w_{2,6}\left(\ln(1 - w_{1,6}\,[I_2 - 3])\right) \\ &+ w_{2,7}w_{1,7}\,[I_1 - 3]^2 + w_{2,8}\left(\exp(w_{1,8}\,[I_1 - 3]^2) - 1\right) - w_{2,9}\left(\ln(1 - w_{1,9}\,[I_1 - 3]^2)\right) \\ &+ w_{2,10}w_{1,10}[I_2 - 3]^2 + w_{2,11}\left(\exp(w_{1,11}[I_2 - 3]^2) - 1\right) - w_{2,12}\left(\ln(1 - w_{1,12}[I_2 - 3]^2)\right)\end{aligned} \tag{20}$$

In a manner similar to multiple regression, this energy function comprises 12 terms embedded with 24 unknown weightings from the input and hidden layer. To reduce redundancy, the first two coefficients before the linear terms can be combined into one, resulting in a total of 20 unknown weighting estimates, the same as in multiple regression. These weightings are learned by neural network by minimizing the loss function, which is also based on mean absolute percentage error. However, we employed a different approach to mitigate potential overfitting by regularizing or shrinking the weighting estimates towards zero.

$$\mathcal{L}(b; F) = \frac{100}{N} \sum_{i=1}^{N} \left\| \frac{P(F_i) - \widehat{P}_i}{P(F_i)} \right\|_1 + \alpha_1 \|w\|_1 + \alpha_2 \|w\|_2^2 \qquad (21)$$

$\alpha_1 \|w\|_1$ and $\alpha_2 \|w\|_2^2$ refer to LASSO and Ridge regularization, respectively. Here, $\|w\|_1$ and $\|w\|_2$ denote the L1 and L2 norm of weighting coefficient, $\alpha_1$ and $\alpha_2$ are the corresponding regularization hyperparameters. The combination of LASSO and Ridge regularization is commonly recognized as the elastic net. In this study, we learned the weighting coefficients by minimizing the loss function, as defined in Eq. 21, with the aid of gradient-based adaptive optimizer (ADAM), while simultaneously enforcing a non-negative constraint on all weightings.

**2.4 Data preparation and implementation details**

In order to compare the predictive capacity of multiple regression and neural network, we utilized the stress-strain data from the human brain cortex, including uniaxial tension, compression, and simple shear, for both model regression and training processes. The data was obtained from literature [21]. To investigate the effectiveness of our proposed framework, we considered two regression and training scenarios: single-mode and multi-mode. In the single-mode scenario, we employed a single loading case, either tension, compression, or shear, as the regression or training data, while the remaining two were used as test data. On the other hand, in the multi-mode scenario, we simultaneously regressed or trained our model with data from all three loading cases. For the multi-mode scenarios, the objective loss function was obtained as a linear combination of the loss function for each loading case, i.e., $\mathcal{L}^{ALL} = \mathcal{L}^{Ten} + \mathcal{L}^{Com} + \mathcal{L}^{Shr}$. We conducted the training process on a Legion PC equipped with a six-core Intel Core I7-8750H 2.2GHz CPU, 4 GB NVIDIA GTX 1050Ti GPU, and 24GB of memory, while the neural network source code was primarily obtained from literature [43].

# 3 Results

## 3.1 Model calibration

Prior to implementation, the model for strain energy function, Eq. (12), has to be calibrated to determine the optimal choice of its hyperparameters. In the case of multiple regression, the hyperparameter pertains to the number of terms in Eq. (17) used to regress the model, while for neural network, the regularization penalty parameters, i.e., $\alpha_1$ and $\alpha_2$, in Eq. (21) serve as the hyperparameter.

### 3.1.1 Model terms selection for multiple regression

Figure 2 illustrates the BIC value obtained from multiple regression models with respect to different numbers of terms in Eq. (17), where color-filled regions evaluate the contributions of each term from a set of twelve terms, as shown at the bottom of the figure. Four different training scenarios have been considered, including training individually with the stress-strain data collected from either tension, compression or shear experiments, and training simultaneously with data from all three loading modes. As depicted in Figure 2, the lowest BIC value corresponds to a model with two terms for all the single-mode training scenarios, and a single-term model for the multi-mode training scenario. The BIC measures the trade-off between model accuracy and penalty for model complexity, indicating that the optimal model should be accurate enough while being as simple as possible. This is clearly demonstrated in Figure 2b, where the optimal model emerges with two terms based on the lowest BIC. If we introduce more terms during regression, such as four terms, the accuracy remains nearly the same since the selected terms $[I_1 - 3]^2$ and $[I_2 - 3]$ continue to be dominant, while additional terms contribute trivially. However, the selected model tends to be more complex. In an extreme case, calibrating the model with 12 terms results in a final model with eight terms. Although the model seems to be more accurate than the two-term model for the current dataset, it is more likely to lose its generality for other datasets, which is referred to as overfitting. Therefore, we choose the model with two terms as the final model, which has the highest accuracy and interpretability, as well as the simplest form.

**3.1.2 Regularization hyperparameter for artificial neural network**

In order to prevent the neural network from overfitting, we incorporated the regularization effect by introducing two additional terms into the loss function, as described in Eq. 21, where $\alpha_1$ and $\alpha_2$ are hyperparameters for L1 and L2 regularization, respectively. The combination of L1 and L2 regularization is also commonly referred to as the elastic net regularization.

Figure 3 illustrates the impact of hyperparameter tuning for L1 regularization on the selection of neural network models, where four different regularization parameters including 0, 0.01, 0.1 and 1 were utilized. For each case, the model was trained simultaneously with stress-strain data from three loading modes, namely tension, compression, and shear data for the human brain cortex. Then, the model tested on each loading mode individually to evaluate its predictive performance measured with R-square value. The color-filled regions in the figure represent the contributions of the 12 terms, as shown at the bottom of the figure.

In the absence of regularization ($\alpha_1 = 0$), as shown in Figure 3a, the optimal model consists of five terms, which performs well in predicting all three loading modes, with $R^2_{train}$ values of 0.848, 0.741, and 0.981 for tension, compression, and shear, respectively. When the model is mildly regularized with $\alpha_1 = 0.01$, as depicted in Figure 3b, the model remains accurate, while the number of selected terms decreases to four. This clearly indicates the impact of regularization in reducing the number of identified terms.

In the case of moderate regularization ($\alpha_1 = 0.1$), the number of model terms continuously decreases to two, as shown in Figure 3c. Compared to the non-regularized model, the $R^2_{train}$ slightly increases from 0.848 to 0.857 for tension, while there is a minor decrease of 0.009 and 0.004 for compression and shear, respectively. Despite this mild loss in accuracy, the discovered terms remain the same as those dominant in the non-regularized model and mild-regularized model, i.e., $[I_2 - 3]$ and $-\ln(1 - [I_2 - 3]^2)$. When the model is substantially regularized with $\alpha_1 = 1$, as illustrated in Figure 3d, the discovered model still consists of two terms. However, the fit quality decreases significantly for all three loading modes, especially for the compression, where $R^2_{train}$ drops from 0.741 in the non-regularized model to 0.656 in the heavily regularized model. Moreover, the model discovers a new term $[I_1 - 3]^2$, which is not present in the previous regularized models. These findings suggest that the model is over-regularized. Therefore, the neural network model regularized with $\alpha_1 = 0.1$ is capable of obtaining the simplest form with satisfying accuracy. Analogously, we also examine the effect of L2 regularization hyperparameter tuning on neural network model selection, as illustrated in Figure 4, using four different regularization parameters (0, 0.01, 0.1 and 1). Interestingly, even for a substantially regularized model with $\alpha_2 = 1$, the number of discovered terms remain the same. This finding is in contrast to Linka et al. [45], where a L2 regularization with $\alpha_2 = 0.001$ is introduced to reduce the number of trivially important terms. The discrepancy may be attributed to the difference in the optimization metric used in the loss function. The MSE metric was used in their study, while this study used the MAPE to amplify the difference between data points by converting the absolute difference into relative one expressed in percentage. The main advantage of MAPE is to reduce the time for training, while maintaining or even improving the model accuracy, especially for dense data [56]. In our case, the tension experiment data, which exhibits high nonlinearity within a small range of stretch (0.1), can be better represented and evaluated by the MAPE intrinsically. Additionally, L1 regularization, also known as Lasso regression, penalizes unnecessary complexity of the model to obtain sparse sets of features [57]. Conversely, L2 regularization, known as ridge regression, penalizes the contribution of non-dominant terms to make the important term more prominent. Therefore, L1 regularization is widely used in feature selection, whereas L2 regularization is employed to constrain the range of weightings [58-60].

For elastic net regularization, the effect of hyperparameter tuning on model selection is illustrated in Figure 5. The optimal regularization is obtained with $\alpha_1 = \alpha_2 = 0.01$. As the combination of L1 and L2 regularization, the elastic net regularization incorporates the characteristics of both methods. Therefore, as $\alpha_1$ and $\alpha_2$ increase to 0.1, the model tends to discover more terms, as is observed with L2 regularization.

## 3.2 Comparison in model performance between multiple regression and artificial neural network

Figure 6 illustrates the hyperelastic material model for the human brain cortex obtained through multiple regression, with a comparison of single-mode and multi-mode training scenarios. As detailed in Section 3.1.1, during the single-mode training (i.e., tension, compression, shear), the model is regressed using two randomly selected terms from a set of 12 terms, while multi-mode training employs only one term for regression. The figure showcases that the regression models derived from the single-mode training exhibit superior interpolation for each of the three individual training sets, with an excellent goodness of fit ($R^2_{train}$ = 0.999, 1.000, 1.000).

Nonetheless, the performance of the regression models demonstrates a noticeable decline for the remaining two sets of test data. Specifically, the model trained with tension data underestimates both compression and shear test data (Figure 6a), with corresponding $R^2_{test}$ value of 0.439 and 0.938, respectively. Conversely, the model trained with compression data tends to overestimate the prediction of test data, with $R^2_{test}$ value of 0 for tension and 0.647 for shear (Figure 6b). The model trained with shear (Figure 6c) performs moderately in between, overvaluing the tension with a $R^2_{test}$ value of 0.605, but undervaluing the compression with a $R^2_{test}$ value of 0.645. It should be noted that the above estimation for compression is based on its absolute value. For multi-mode training, the goodness of prediction $R^2_{train}$ slightly decreases to 0.869, 0.768 and 0.999 compared to those of single-mode training models, while the collective goodness, i.e., the sum of $R^2$, increases.

Figure 6 also shows several intriguing observations. First, it is evident that the model trained on tension data performs inadequately in predicting compression data, while the model trained on compression data fails to predict tension data, with a $R^2_{test}$ value of 0. This incompatibility highlights the asymmetric feature present in tension and compression for the human brain tissue, including the cortex [21, 61, 62]. Second, a comparison between Figure 6c and Figures 6d reveals a consistent trend in model prediction, suggesting that shear plays a dominant role in multi-mode training. In other words, if only one single experiment is available to calibrate the hyperelastic model of the human brain cortex, shear data would be prioritized. This suggestion is consistent with previous research [63].

The hyperelastic model discovered by neural network is depicted in Figure 7. The overall trends are similar to those observed in the multiple regression. The model trained with a single-mode data, i.e., tension, compression, or shear, performs much better in predicting training data than testing data, as illustrated in Figure 7a -7c. In multi-mode training, the model is less accurate for prediction in each loading mode compared to single-mode training model, but the collective goodness is better. Whereas following the similar trend, the model derived from neural network underperforms in predicting all the shear modes, especially for the model trained individually with tension and compression, where the $R^2_{test}$ decreases from 0.938 to 0.656, and from 0.647 to 0.422, respectively. A mutual comparison is demonstrated in Figure 8, where all the models are trained with three loading modes simultaneously. As revealed, the model derived from multiple regression outperforms that derived from the neural network in predicting all the three loading modes.

Remarkably, the multiple regression model and the neural network model exhibit comparable performance under small deformations. As depicted in Figure 8a, the curves almost coincide with each other for the entire range of tensile strains. For compression in Figure 8b, the curves overlap until the compressive strain reaches 0.075, while for shear in Figure 8c, the departure occurs when the shear strain approaches 0.15. After the departure, the blue curve of the multiple regression model moves closer to the experiment data points, indicating superior performance compared to the neural network model. In summary, the multiple regression model outperforms the neural network model primarily under large deformations, resulting in a higher goodness of fit with a greater $R^2$ value.

### 3.3 Comparison in model terms selection between multiple regression and artificial neural network

In Figure 9, the model terms discovered from multiple regression and neural network under single-mode training and their corresponding contributions are depicted. Different colored regions indicate the contribution of selected terms, with larger areas indicating higher contributions. The results demonstrate that the multiple regression model has significantly fewer terms than the neural network model, while still achieving a higher or comparable goodness of fit for the individual loading modes. For example, the model regressed with tension data consists of two terms (Figure 9a) and achieves better performance than the four-terms model trained by neural network (Figure 9d). The two-term regression model for compression (Figure 9b) obtains comparably perfect goodness of fit (i.e., $R^2_{train} = 1.000$) as the five-term neural network model (Figure 9e). Similarly, the two-term regression model for shear data (Figure 9c) outperforms the six-term neural network model (Figure 9f), with a $R^2_{train}$ value increasing from 0.985 to 1.000. The detailed weightings of both models are summarized in Table 1.

A closer examination of the discovered terms shows that the dominant terms discovered by multiple regression are also included in the terms selected by the neural network model, particularly for compression and shear data. This indicates that the neural network model incorporates additional insignificant terms that contribute marginally to the model performance. As elaborated in Section 3.1, the ideal model should achieve the highest accuracy in prediction with the simplest possible form. Hence, the model trained by neural network is not optimal. Furthermore, the contributions of selected terms provide insight into the strain-stiffening effect of the human brain cortex [21, 64, 65]. For example, in Figure 9b, the model can accurately predict the stress-strain relations under small deformations with a single linear term $[I_2 - 3]$. However, for large deformations, the curve exhibits pronounced nonlinearity, leading to the failure of the linear term. This implies that a nonlinear term (i.e., $[I_1 - 3]^2$), is necessary to capture the intrinsic nonlinearity of the human brain cortex under large compressive deformations.

Figure 10 illustrates the model terms discovered through multiple regression and neural network under multi-mode training scenario. Notably, the model derived from multi-mode training exhibits a tendency to select fewer terms than models trained with single mode, particularly in the case of the neural network model, which comprises only two terms compared to five or six terms selected in single-mode training. Again, the multiple regression model outperforms the neural network model in prediction for all three loading modes, with a relatively simpler form. For example, the one-term regression model accurately predicts the shear data (Figure 10c), with a $R^2_{train}$ value of 0.999, while the two-term neural network model achieves only 0.977 in $R^2_{train}$ value (Figure 10f). The multiple regression model incorporates linear exponentials of the second invariants ($e^{[I_2-3]} - 1$), which bears a resemblance to the Demiray model that employs the first invariant as the variable (see Eq. 15). On the other hand, the neural network model is a linear combination of two terms, $[I_2 - 3]$ and $-\ln(1 - [I_2 - 3]^2)$, which is a combined form of the Mooney-Rivlin model and the Gent model. Strikingly, both models are a function of the second invariant $I_2$, and this is consistent with previous study [66].

It is essential to note that aforementioned models were established on the basis of fixed criteria. For example, the optimal regression model was determined by the lowest BIC value, while the resultant neural network model was regularized using Lasso regression with a hyper-parameter $\alpha_1$ of 0.1. In practice, it is expected that various models will be obtained with different criteria. Figure 11 shows the top two models derived from the multiple regression and neural network, respectively, where all models were trained using three loading mode data simultaneously. As depicted, the one-term model, which corresponds to the lowest BIC value (Figure 11a), exhibits consistent performance with the two-term model obtained based on the highest adjusted R-squared value (Figure 11b). Conversely, the two-term model regularized with Lasso regression (Figure 11c) outperforms the three-term model regularized with elastic net (Figure 11d).

### 3.4 Comparison in model performance with classic models

Figure 12 compares the performance of the models derived from multiple regression and neural network with classic invariant-based models, such as neo-Hookean, Mooney Rivlin, Demiray and Gent. In the case of single-mode training scenarios (Figure 12 a-c), the Demiray, Gent and multiple regression models accurately agree with the experiment data, while the neural network model fails to capture the nonlinearity in stress-strain relationships under large deformations. On the other hand, the neo-Hookean model and two-term Mooney Rivlin model, which are predominantly composed of linear terms, are inadequate in representing the experimental data. In multi-mode training scenarios (Figure 12 d-f), although most models perform worse compared to those derived from single-mode training, the multiple regression model and neural network model still perform relatively better. To be precise, all models overestimate the tensile experimental data (Figure 12d), while the models derived from multiple regression and neural network move closer to the experimental data. For compression data, all models have equal performance in underestimating the absolute value of compressive stress (Figure 12e). In contrast, the regression model perfectly represents the shear data (Figure 12f), while the neural network model moderately undervalues the shear stress under large deformations. However, the classic models, except for the linear-form neo-Hookean model and Mooney Rivlin model, all overvalue the shear data. Overall, our results demonstrate the superior performance of multiple regression and neural network models in accurately capturing the stress-strain relationships, especially in multi-mode training scenarios.

## 4 Discussion

As the most structurally and functionally intricate organ, the human brain possesses highly complex mechanical properties, such as heterogeneity, ultra-softness, and biphasic behavior [1]. Accurately characterizing these properties using a constitutive model is essential for understanding traumatic brain injury [10, 11], brain development [3, 4, 67] and disease progression [6, 7]. Recently, Linka et al. proposed a method using constitutive artificial neural network to discover optimal constitutive models automatically. This method satisfies several physical priors, including thermodynamic consistency, material objectivity, polyconvexity and physical constrains, and has been successfully implemented in discovering the constitutive models for rabbit skin [44] and human brain tissues [45]. However, the discovered model for human brain tissue is not concise in format, particularly for the model derived from single-mode training, which consists of over ten terms. The complicated form limits its generality and raises concerns about overfitting by increasing the variance in prediction. Additionally, the neural network's black box nature hampers its interpretation and understanding. To address these limitations, we employed traditional statistical methods, specifically multiple regression, to discover the constitutive model for the human brain tissue following the same steps as neural network. By conducting a comprehensive comparison between these two methods, we seek to answer the following questions: how does the neural network work from a statistical perspective? Is the model discovered by the neural network globally optimum for given options? And if not, how can we find them?

*The artificial neural network and multiple regression are fundamentally equivalent.* Following preliminary studies on various loss functions such as mean squared error (MSE), mean absolute error (MAE), and mean absolute percentage error (MAPE), we ultimately selected MAPE as the final objective function. By comparing Eq. 19 and Eq. 21, it is evident that both methods share similar objective functions, with the only difference being the regularization terms. To prevent potential overfitting, the neural network method employs Lasso regression (L1), Ridge regression (L2), or Elastic net regularization (L1 and L2), while multiple regression uses information criteria such BIC to balance the trade-off between model fitting accuracy and complexity. From a statistical perspective, both methods fall under the multivariate analysis and are statistically equivalent despite the different approaches to penalize model complexity. In addition, the best subset selection algorithm is utilized in multiple regression to traverse the model selections, where we access and compare all possible models including single term to 12 terms, and the optimal model corresponds to the lowest BIC value. However, in neural network, we fit the model with all 12 terms and employ the regularization terms to shrink the coefficient of unimportant predictors towards zero.

*The artificial neural network can automatically discover accurate constitutive models for human brain cortex.* By adhering to constitutive theories and thermodynamic constraints, the neural network is capable of identifying a series of constitutive models that accurately reflect the experimental data. In the single-mode training, the neural network discovered four-term, five-term, and six-term models with remarkable fitting performances of 0.964, 1.000 and 0.985 for tension, compression, and shear, respectively. In contrast, the multi-mode trained models comprise only two terms but achieve respective accuracy values of 0.857, 0.732, 0.977. Furthermore, the neural network results exhibit some intriguing characteristics as noted in previous studies [45, 61, 63]. First, the discrepancy in predicative performance for model trained individually on tension and compression data reveals the pronounced asymmetric nature of tension and compression in the human brain tissue. Second, the consistent trend in the prediction of models trained on shear data and multi-mode data suggests that shear plays a dominant role in multi-mode training scenario, which indicates that shear experiment should be the primary choice if only one experiment is feasible due to the limited availability of brain tissue specimens. Last, all terms selected by multi-mode training are associated with the second invariant $I_2$. Compared to multiple regression, a predominant advantage of neural networks is its ability to automate mode discovery. While multiple regression requires manual traversal of all variables, the neural network can perform this task simultaneously, significantly reducing the time required for searching the optimal model, particularly when dealing with enormous amounts of available terms.

*The model discovered by the artificial neural network is not globally optimal and could be further simplified using multiple regression.* In the case of single-mode training, the multiple regression model achieves higher accuracy with only two terms, compared to the neural network model that requires at least four terms. In multi-mode training, both methods tend to select fewer terms, but the single-term multiple regression model still obtains comparable accuracy compared to the two-term neural network model. These findings suggest that, in the current scenario, the multiple regression model performs better than the neural network model in terms of accuracy and simplicity, indicating that the neural network model may not be globally optimal. However, it is important to note that the comparison between the two models is limited by the fact that cross-validation was only performed using four different values for $\alpha_1$. The disparity between the two models suggests that L1 regularization with $\alpha_1 = 0.1$ may not be the optimal set of

hyperparameters for the neural network. To address this issue, it is recommended to perform hyperparameter tuning for each training scenario using cross-validation and incorporating more options for $\alpha_1$ and $\alpha_2$. However, this approach may increase the time required for hyperparameter. Thus, careful consideration of the trade-off between computational cost and performance improvement is necessary.

*Both artificial neural networks and multiple regression methods offer flexibility in model discovery with distinct metrics*. The neural network has the ability to discover a variety of models based on various sets of hyperparameters. Moreover, distinct weight initialization can result in significant different neural network models, particularly in the context of single-mode training. Similarly, the optimal multiple regression model tends to yield diverse structures when evaluated with different metrics, such as Akaike information criterion (AIC), corrected Akaike information criterion (AICc), R-squared, and adjusted R-squared. Consequently, several models with comparable performance in data fitting can be obtained, indicating that the model is non-unique based on the current experimental data. Incorporating additional experimental data or more loading modes may help in obtaining a unique model. On the other hand, this non-uniqueness also suggests flexibility in model discovery with distinct metrics, thereby enriching the selection of optimal constitutive models.

*Limitations and future application.* In the present study, we employed the best subset selection algorithm for multiple regression analysis. While effective, this approach may become computationally infeasible as the number of terms increases ($2^n - 1$). To overcome this issue, we suggest using the stepwise selection method, either forward or backward, for model traversal. Additionally, the current approach to enforce polyconvexity via additive strain energy functions is strictly sufficient, but other options such as multiplicative coupling between individual invariant (such as $[I_1 - 3] \cdot [I_2 - 3]$ term in the three-term Mooney Rivlin model [68, 69]) could also satisfy the polyconvexity requirement and expand model selection. However, additional constraints should be considered to ensure polyconvexity, such as the positive semi-definite Hessian matrix of the strain energy function [36]. Finally, while our models are based on invariants, principal-stretch-based models have shown better capability in capturing the characteristic behavior of stress-strain relations under combined loading and strain-stiffening [28, 63, 70]. Therefore, exploring our approach within the framework of principal stretch would be an intriguing application for future research.

# 5 Conclusion

Human brain tissue is one of the most important and complex tissues in the body. Understanding the mechanical behavior of the brain tissue will greatly contribute to unwrapping the underlying mechanisms of brain development and disease progression. Despite numerous constitutive models proposed for the brain tissue, due to its complex mechanical properties including ultrasoftness, heterogeneity, biphasic and highly fragile behaviors, a prevailing consensus has not been reached. To overcome this challenge, we employed both artificial neural network and multiple regression methods to discover the optimal constitutive material model for the human brain cortex. To ensure the comparability of both methods, we kept all setups consistent across both methods, except for the distinct approach to prevent potential overfitting. The optimal model for multiple regression was determined with the lowest BIC value, which balanced the trade-off between model accuracy and complexity. Our results demonstrate that the multiple regression model tends to select fewer terms compared to the neural network, while achieving higher prediction accuracy. This is particularly noticeable when the models are trained using single-mode training, where two-term regression model obtains better performance than the six-term neural network model in predicting the shear data, with the $R^2$ value improving from 0.985 to 1.000. Moreover, our findings highlight the importance of the regularization hyperparameters for the neural network. With more detailed cross-validation in $\alpha_1$ and $\alpha_2$, it is foreseeable that the neural network model can achieve comparable performance to the multiple regression. However, the time required in hyperparameter would increase accordingly, as they should be performed individually for each training scenario. With proper regularization, the artificial neural network has the promising potential to automatically discovering accurate constitutive models that can capture the sophisticated mechanical behaviors of human brain tissue, benefiting computational investigations in traumatic brain injury, brain development and disease progression.

*Acknowledgement*

XW acknowledges the support from National Science Foundation (IIS-2011369); MJR acknowledges the support from National Science Foundation (CMMI-2123061)

*Additional Information*

Competing financial interests: The authors declare no competing financial interests.

*Authorship Contribution Statement*

JH: Methodology, Software, Validation, Investigation, Writing – Original Draft; NF: Validation, Writing – Review and Editing; XC: Methodology, Writing – Review and Editing; JR: Conceptualization, Supervision, Funding acquisition, Writing – Review and Editing; TL: Methodology, Validation, Resource, Funding acquisition, Writing – Review and Editing; XW: Conceptualization, Methodology, Supervision, Funding acquisition, Writing – Review and Editing.

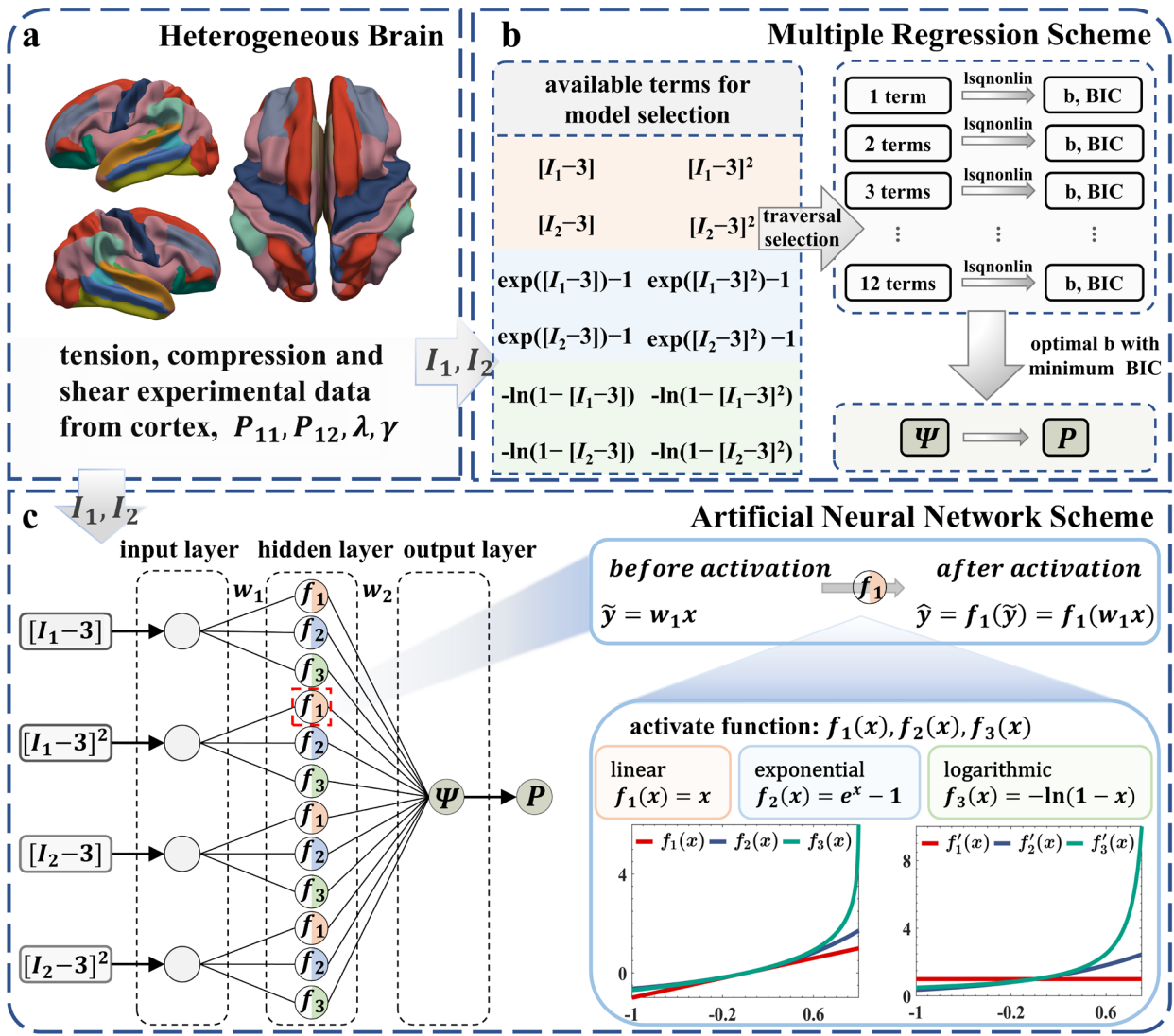

**Fig. 1 Schematic diagram the multiple regression and neural network algorithm utilized for hyperplastic model selections.** (a) heterogeneous brain with anatomical regions rendered in distinct colors, and stress-strain data collected in tension, compression, and simple shear from cortex. (b) The scheme of multiple regression algorithm. (c) The scheme of constitutive artificial neural network

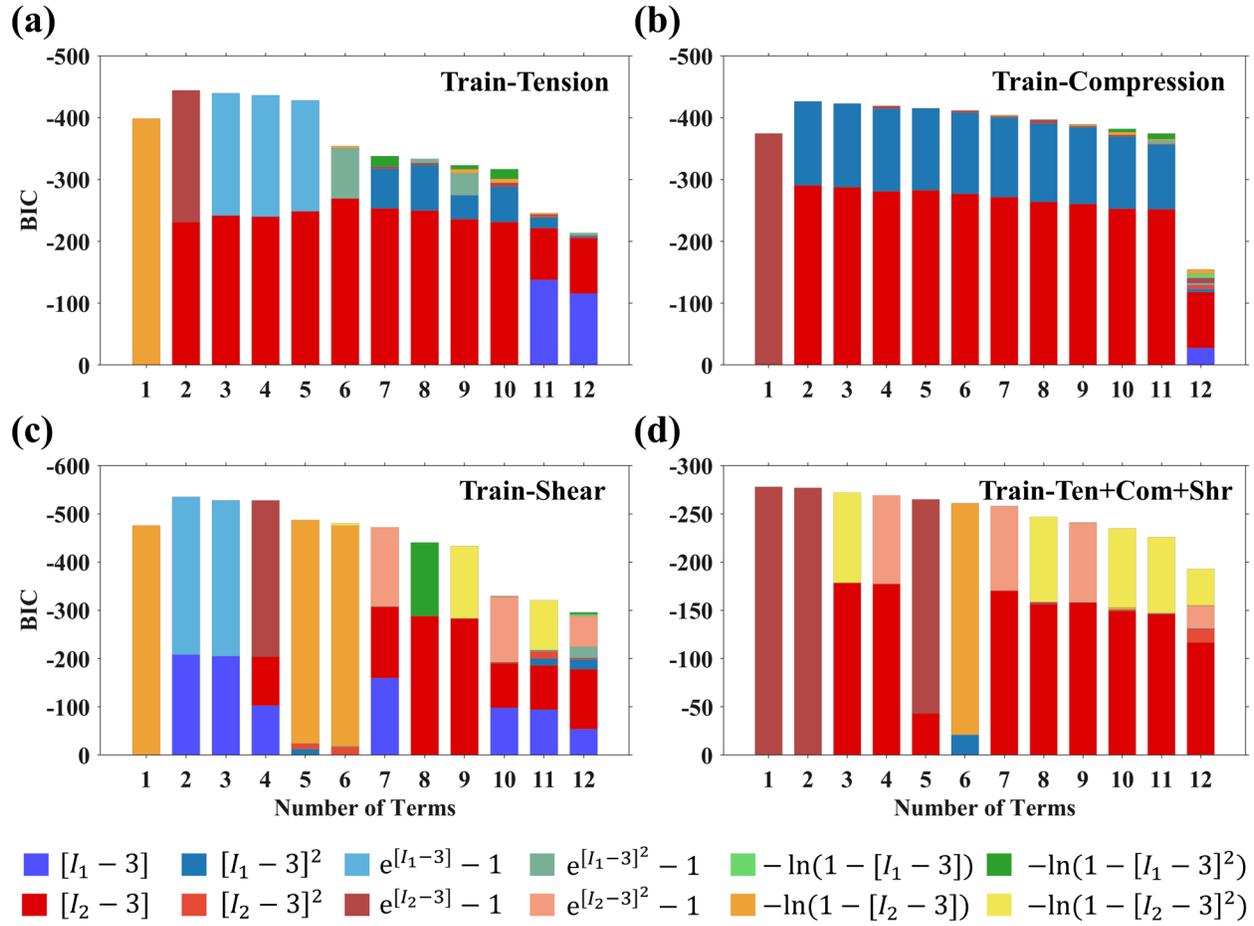

**Fig. 2 Optimal terms determination for multiple regression model.** BIC value obtained through multiple regression with varying number of terms. Training individually using tension (a), compression (b) and shear data (c) from human brain cortex, as well as with all the three loading modes simultaneously (d). Color-filled regions depict the contributions of 12 terms, as illustrated at the bottom of the Figure. Two terms for all the single-mode training scenarios, and single term for the multi-mode training scenario are chosen for multiple regression

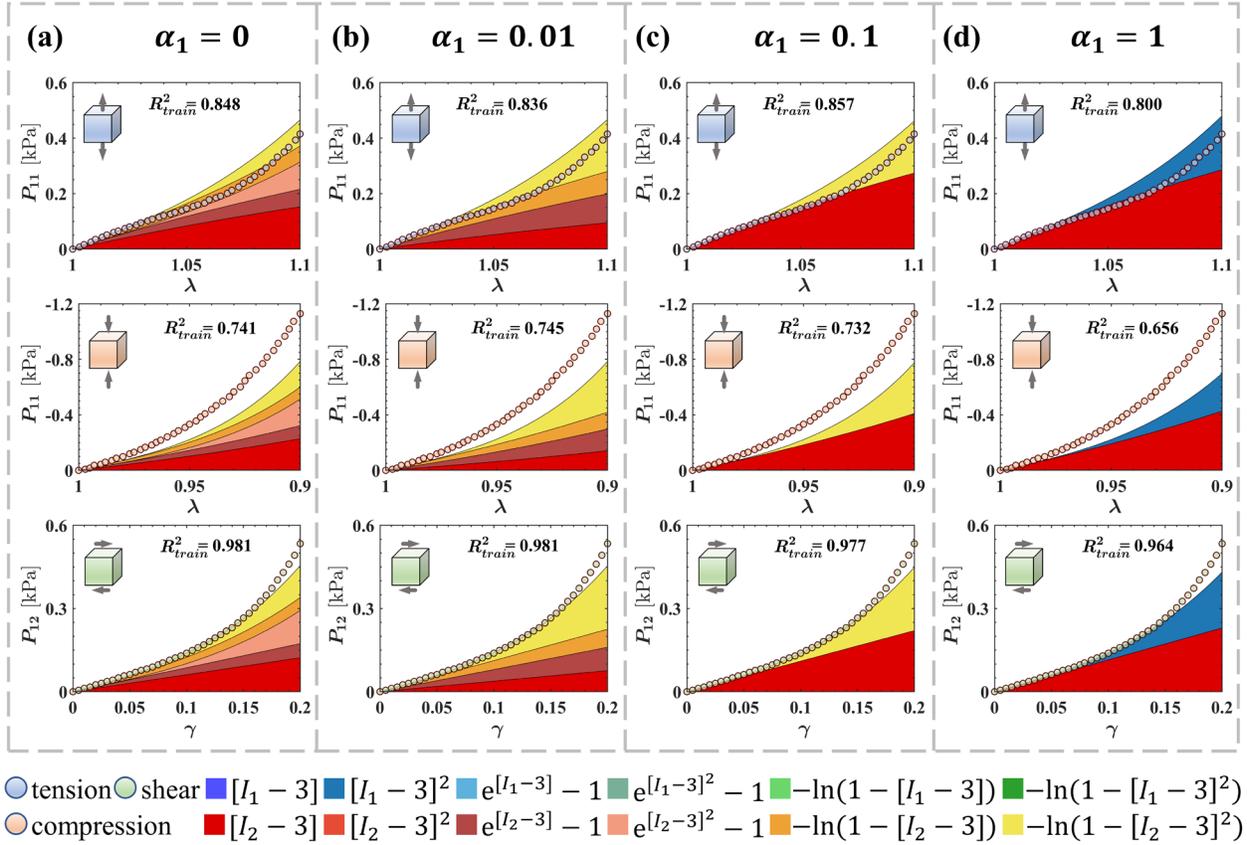

**Fig. 3 Hyperparameter tuning for L1 regularization.** Effect of L1 regularization on model selection for varying penalty hyperparameters. (a)-(d) L1 regularization hyperparameter $\alpha_1 = 0, 0.01, 0.1, 1$. Models are derived from neural networks with multi-mode training, namely trained simultaneously with tension, compression, and shear data. Dots illustrate the experiment data of human brain cortex [21]. Color-filled regions depict the contributions of 12 terms, as illustrated at the bottom of the figure. $R^2$ indicates the goodness of fit. neural network model regularized with $\alpha_1 = 0.1$ is capable of obtaining the simplest form with satisfying accuracy

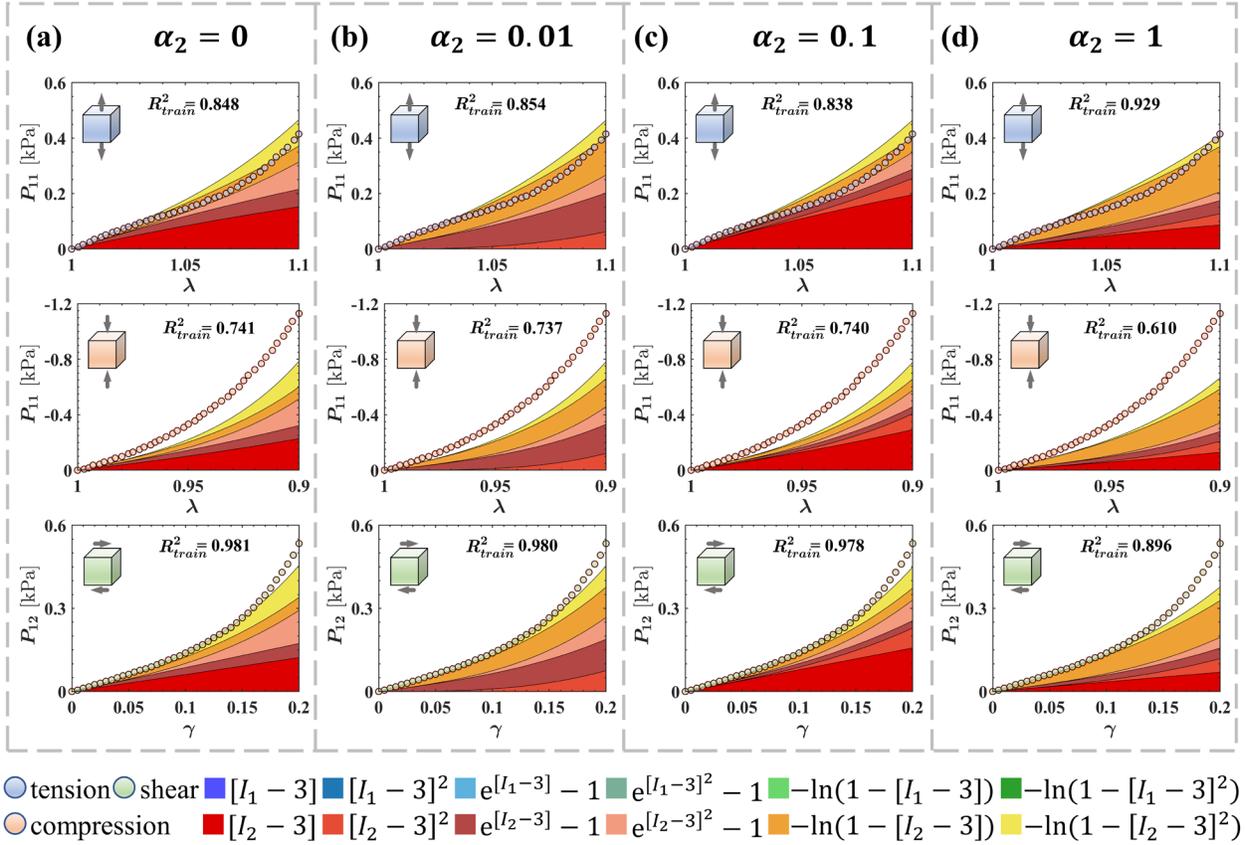

**Fig. 4 Hyperparameter tuning for L2 regularization.** Effect of L2 regularization on model selection for varying penalty hyperparameters. (a)-(d) L2 regularization hyperparameter $\alpha_2 = 0, 0.01, 0.1, 1$. Models are derived from neural networks with multi-mode training, namely trained simultaneously with tension, compression, and shear data. Dots illustrate the experiment data of the human brain cortex [21]. Color-filled regions depict the contributions of 12 terms, as illustrated at the bottom of the figure. $R^2$ indicates the goodness of fit. L2 regularization fails in reducing the number of trivially important terms

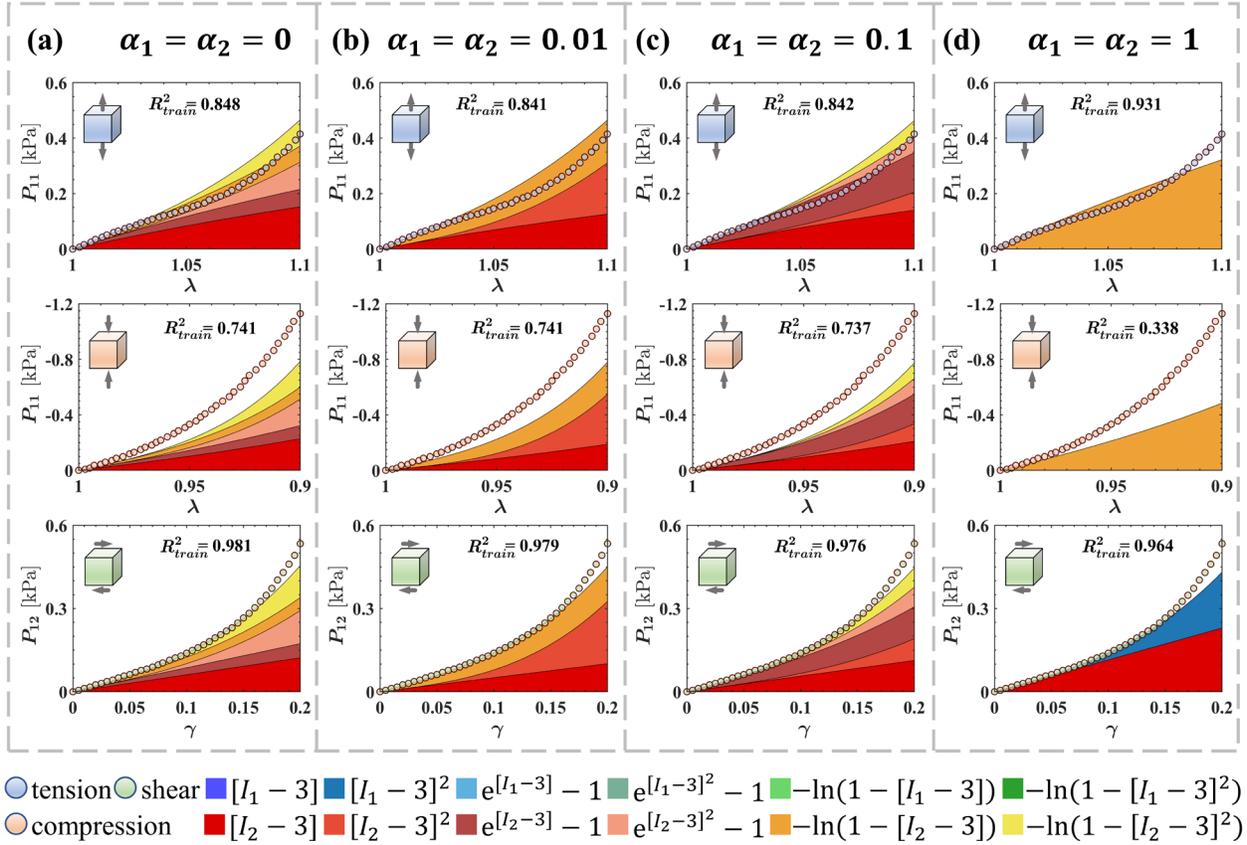

**Fig. 5 Hyperparameter tuning for elastic net regularization.** Effect of elastic net regularization on model selection for varying penalty hyperparameters. (a)-(d) elastic net regularization hyperparameter $\alpha_1 = \alpha_2 = 0, 0.01, 0.1, 1$. Models are derived from neural networks with multi-mode training, namely trained simultaneously with tension, compression, and shear data. Dots illustrate the experiment data of the human brain cortex [21]. Color-filled regions depict the contributions of twelve terms, as illustrated at the bottom of the figure. $R^2$ indicates the goodness of fit. The elastic net regularization incorporates both the characteristics of L1 and L2 regularization

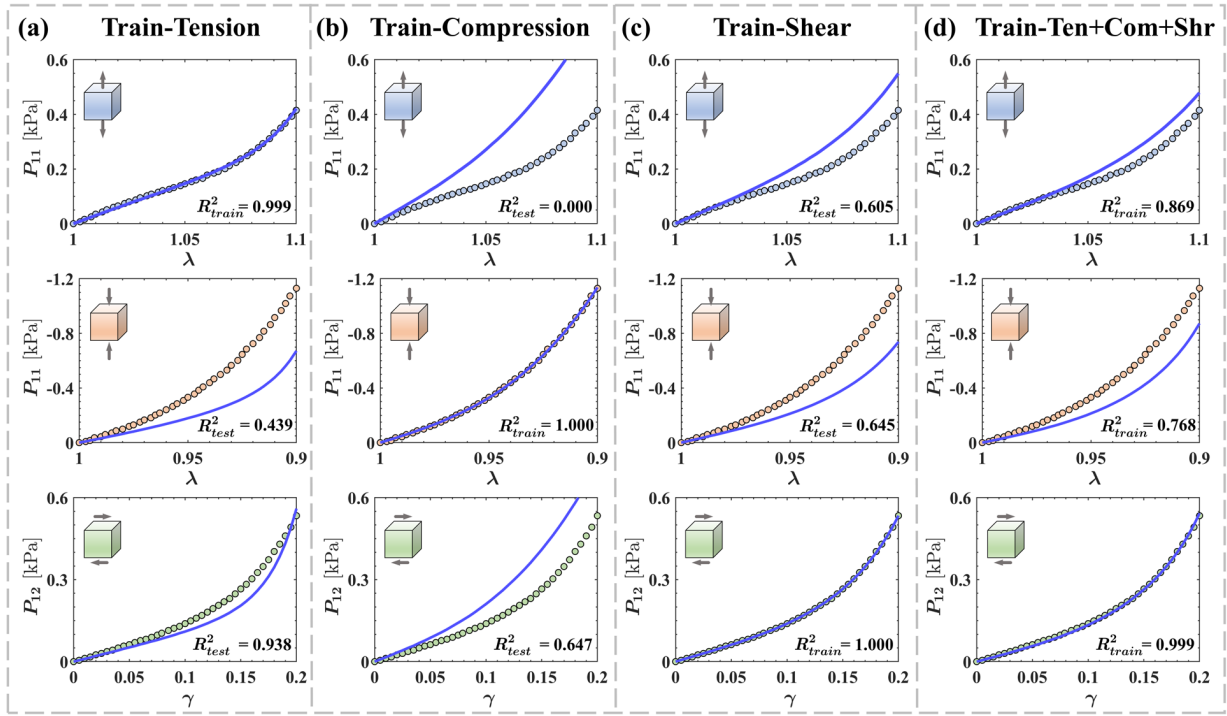

**Fig. 6 Model from multiple nonlinear regression.** Nominal stress as a function of stretch or shear strain derived from multiple regression. Single-mode training with tension (a), compression (b), shear data (c) individually, and testing with remaining two modes. Multi-mode training and testing with data from three loading modes simultaneously (d). Dots illustrate the experiment data of the human brain cortex [21]. $R^2$ indicates the goodness of fit

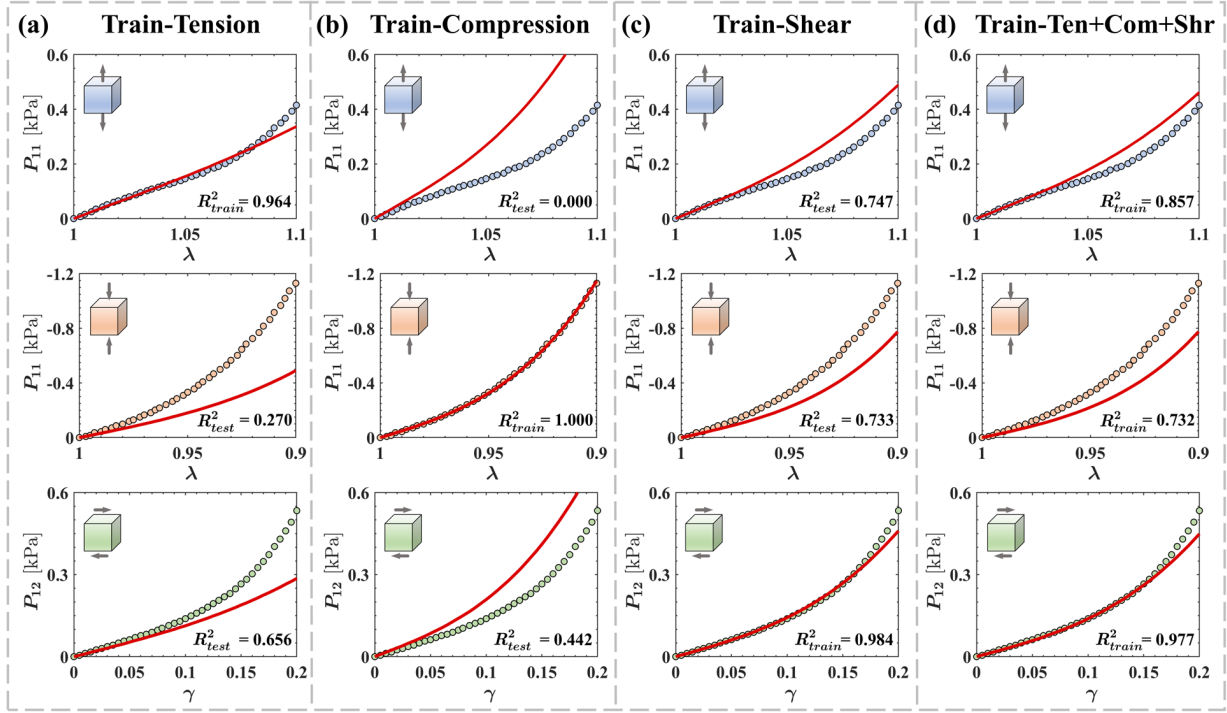

**Fig. 7 Model from artificial neural network.** Nominal stress as a function of stretch or shear strain derived from neural network. Single-mode training with tension (a), compression (b), shear data (c) individually, and testing with remaining two modes. Multi-mode training and testing with data from three loading modes simultaneously (d). Dots illustrate the experiment data of the human brain cortex [21]. $R^2$ indicates the goodness of fit

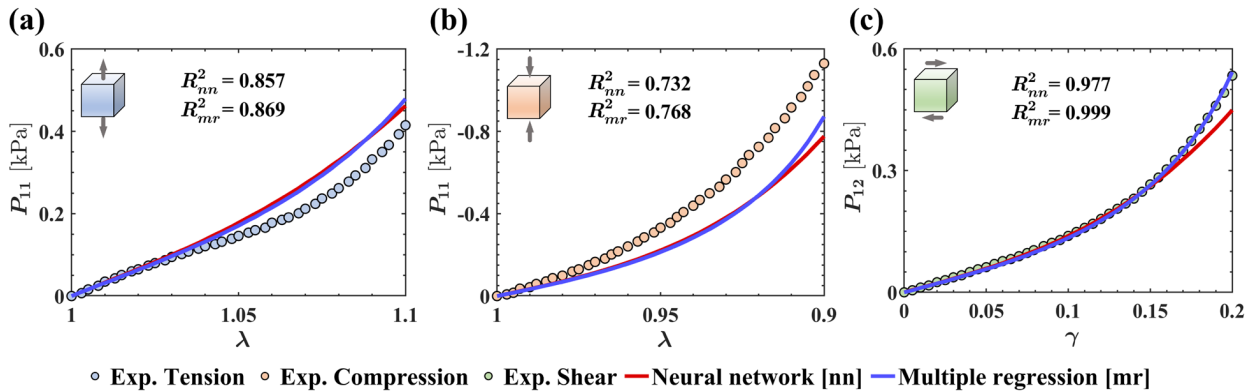

**Fig. 8 Multiple regression vs artificial neural network.** Comparison between predictive performance of models derived from multiple regression [mr] and neural network [nn]. Models are trained simultaneously with data from three loading modes, and tested with tension (a), compression (b) and shear data (c) individually. Dots illustrate the experiment data of the human brain cortex [21]. $R^2$ indicates the goodness of fit. Multiple regression model outperforms the neural network model, particularly under large deformations

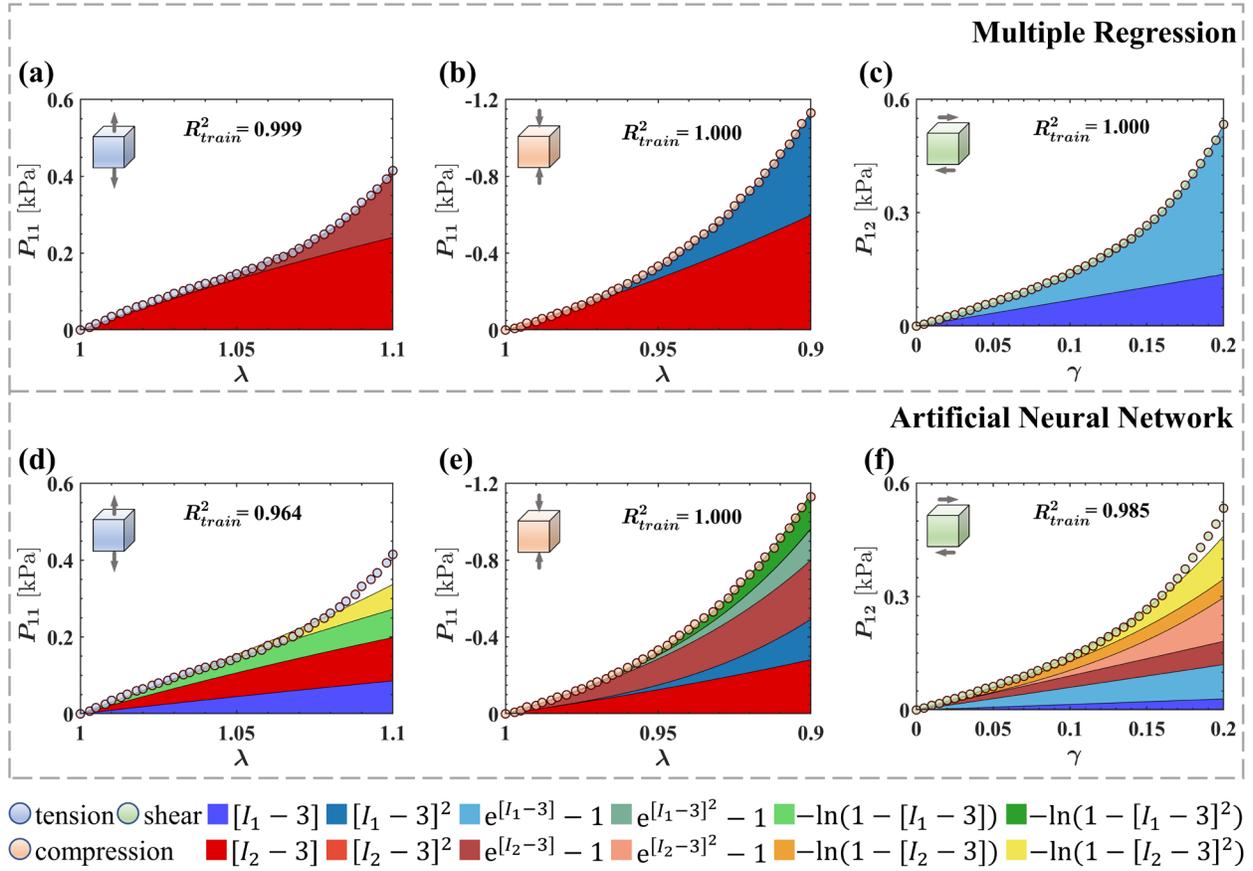

**Fig. 9 Multiple regression vs artificial neural network in model terms discovery under single-mode training.** Comparison in model terms selection between multiple regression and neural network. Models are trained and tested with tension, compression, and shear data individually. (a)-(c) model derived from multiple regression; (d)-(f) model derived from neural network. Dots illustrate the experiment data of the human brain cortex [21]. Color-filled regions depict the contributions of 12 terms, as illustrated at the bottom of the figure. $R^2$ indicates the goodness of fit. The multiple regression model achieves more accurate predictions with much simpler forms compared to the neural networks model under single-mode training

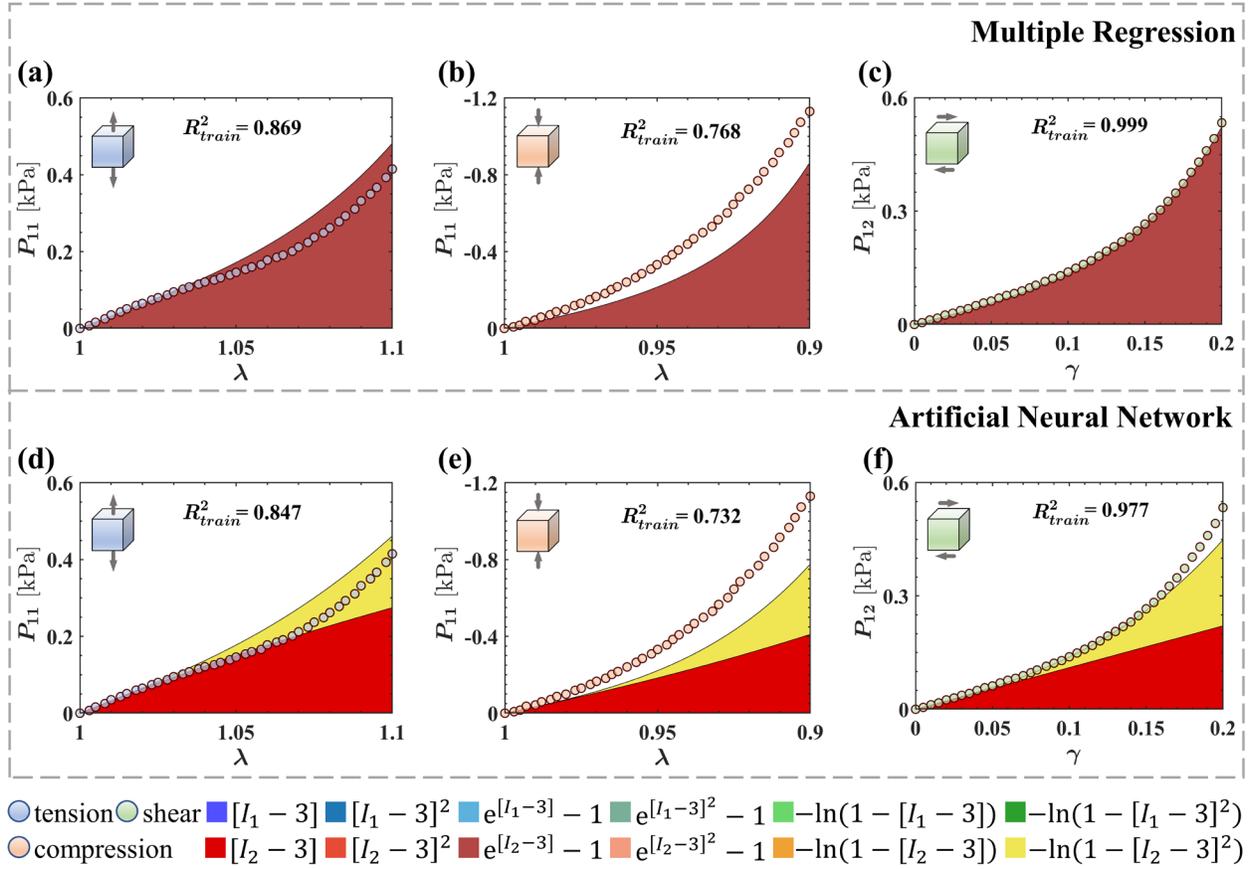

**Fig. 10 Multiple regression vs artificial neural network in model terms selection under multi-mode training.** Comparison in model terms selection between multiple regression and neural network. Models are trained simultaneously with data from three loading modes, and tested with tension, compression, shear data. (a)-(c) model derived from multiple regression; (d)-(f) model derived from neural network. Dots illustrate the experiment data of the human brain cortex [21]. Color-filled regions depict the contributions of twelve terms, as illustrated at the bottom of the figure. $R^2$ indicates the goodness of fit. The multiple regression model obtains comparable accuracy with fewer terms discovered compared to the neural networks model under multi-mode training

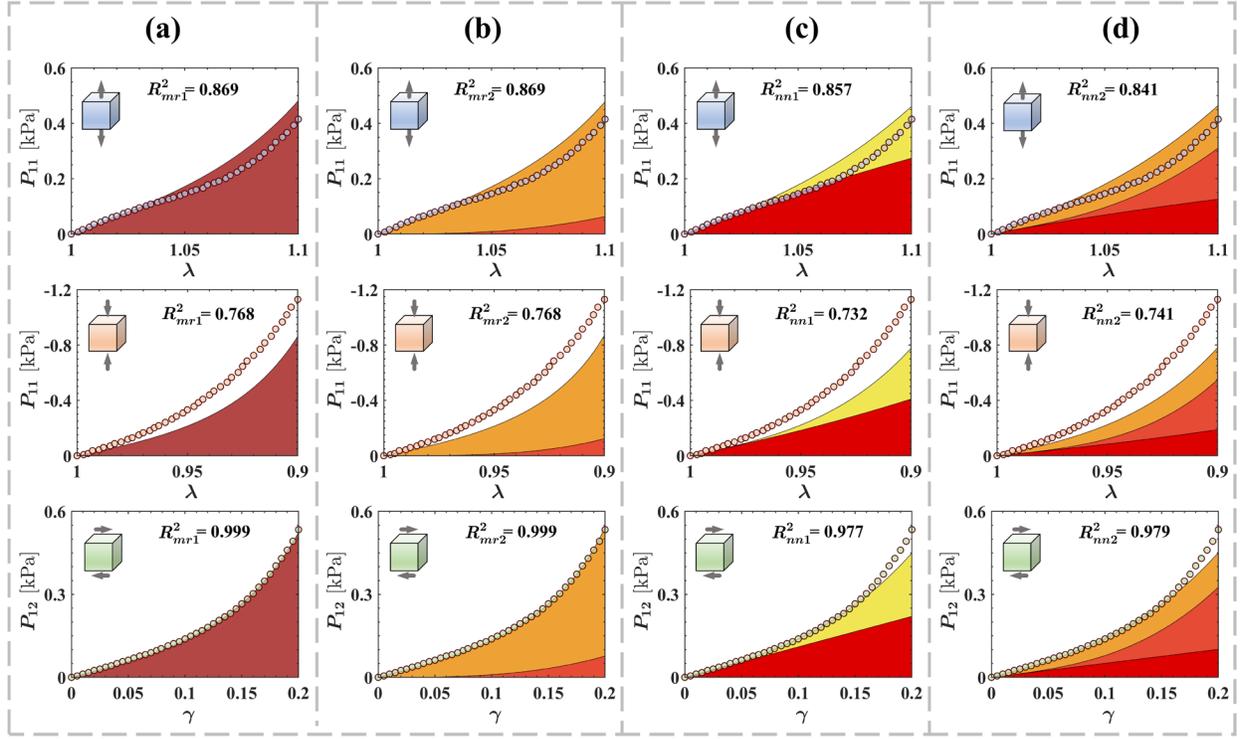

**Fig. 11 Best two models derived from multiple regression vs artificial neural work.** Comparison in model terms and performance between the best two models discovered from multiple regression [mr] and neural network [nn]. (a)-(b) the best and second-best model derived from multiple regression based on BIC and adjusted R-squared, respectively. (c)-(d) the best and second-best model derived from neural network based on L1 regularization ($\alpha_1 = 0.1$) and elastic net regularization ($\alpha_1 = 0.01, \alpha_2 = 0.01$). Dots illustrate the experiment data of the human brain cortex [21]. Color-filled regions depict the contributions of twelve terms, as illustrated at the bottom of the figure. $R^2$ indicates the goodness of fit

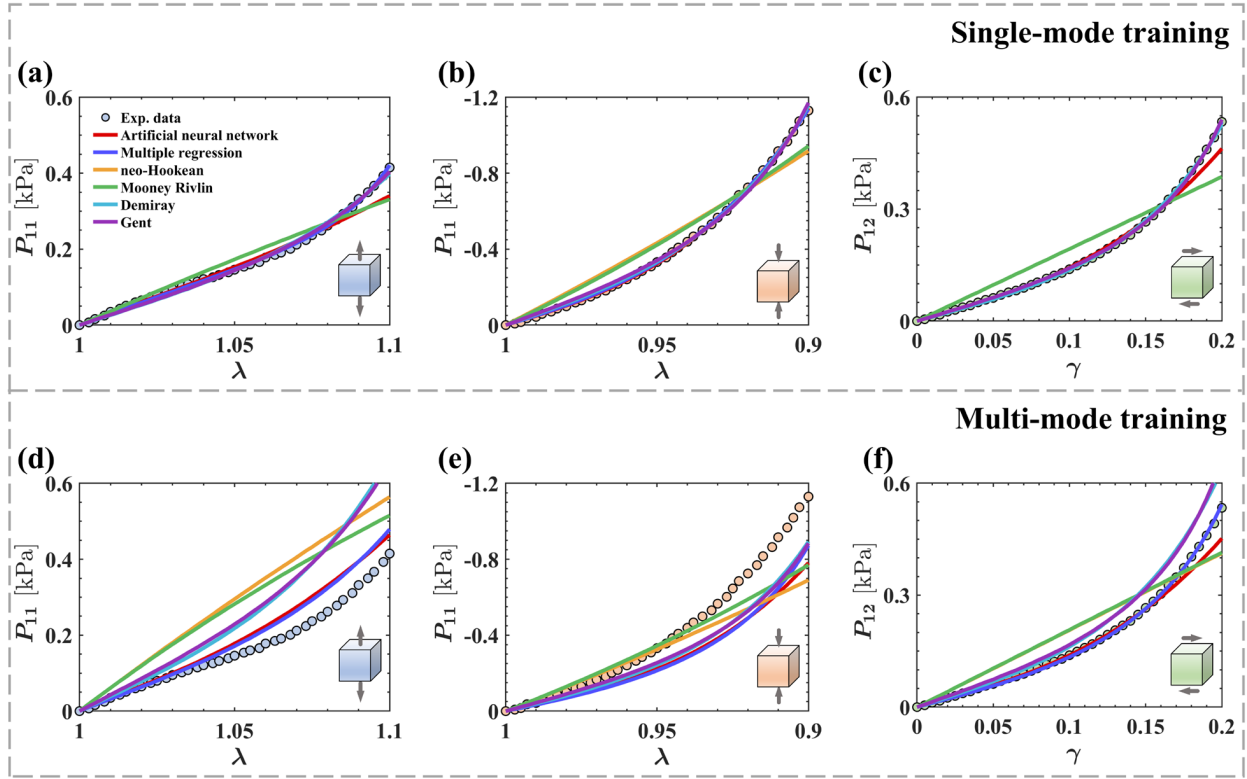

**Fig. 12 Multiple regression, artificial neural network vs classic models.** Comparison between predictive performance of models derived from multiple regression and neural network, as well as those classic models (neo-Hookean, Mooney Rivlin, Demiray and Gent). (a)-(c) models derived from single-mode training, namely trained with either tension, compression, or shear data individually. (d)-(f) models derived from multi-mode training, namely trained with tension, compression, and shear data simultaneously. Dots illustrate the experiment data of the human brain cortex [21]

**Tab. 1** Summary of the twenty unknown parameters ($b_1$ to $b_{20}$) and goodness of fit $R^2$ for single-mode and multi-mode training using loading data from tension (Ten), compression (Com), and shear (Shr). The first 12 parameters represent the weightings of 12 terms, while the remaining 8 parameters are coefficients within each activation function. Notably, the neural network (NN) weightings have been transformed to the same format as the multiple regression (MR) model (refer to Eq. 17)

|  | Train-Ten | | Train-Com | | Train-Shr | | Train-Ten+Com+Shr | |
|---|---|---|---|---|---|---|---|---|
|  | MR | NN | MR | NN | MR | NN | MR | NN |
| $b_1$ | - | 0.15679 | - | - | 0.34293 | 0.07307 | - | - |
| $b_2$ | 0.48450 | 0.22710 | 0.80467 | 0.37908 | - | - | - | 0.55227 |
| $b_3$ | - | - | 12.4972 | 4.88740 | - | - | - | - |
| $b_4$ | - | - | - | - | - | - | - | - |
| $b_5$ | - | - | - | - | 0.00730 | 0.91453 | - | - |
| $b_6$ | 0.00029 | - | - | 0.76190 | - | 0.56883 | 0.02443 | - |
| $b_7$ | - | - | - | 1.99603 | - | - | - | - |
| $b_8$ | - | - | - | 0.18796 | - | 2.12376 | - | - |
| $b_9$ | - | 0.81444 | - | - | - | - | - | - |
| $b_{10}$ | - | - | - | - | - | 0.36796 | - | - |
| $b_{11}$ | - | - | - | 2.15744 | - | - | - | - |
| $b_{12}$ | - | 1.70554 | - | 0.18931 | - | 2.02194 | - | 2.85797 |
| $b_{13}$ | 0.03719 | - | 0.06013 | - | 34.3620 | 0.24588 | 0.09283 | - |
| $b_{14}$ | 96.6665 | - | 0.01778 | 0.53322 | 0.06596 | 0.26781 | 22.1110 | - |
| $b_{15}$ | 0.06296 | - | 0.05931 | 1.89175 | 0.02747 | - | 0.00170 | - |
| $b_{16}$ | 0.05553 | - | 0.05933 | 0.18170 | 0.09116 | 1.69588 | 1.49037 | - |
| $b_{17}$ | 0.00285 | 0.16396 | 0.01768 | - | 0.03603 | - | 0.08627 | - |
| $b_{18}$ | 0.09291 | - | 0.09896 | - | 0.02207 | 0.32871 | 0.04843 | - |
| $b_{19}$ | 0.08868 | - | 0.06196 | 1.92900 | 0.03340 | - | 0.08449 | - |
| $b_{20}$ | 0.00307 | 1.44573 | 0.08173 | 0.18280 | 0.07262 | 1.76025 | 0.02094 | 2.48273 |
| $R_t$ | 0.999 | 0.964 | 0.000 | 0.000 | 0.605 | 0.747 | 0.869 | 0.857 |
| $R_c$ | 0.439 | 0.270 | 1.000 | 1.000 | 0.645 | 0.733 | 0.768 | 0.732 |
| $R_s$ | 0.938 | 0.656 | 0.647 | 0.442 | 1.000 | 0.984 | 0.999 | 0.977 |